\documentclass{article}

\usepackage[preprint]{neurips_2026}

\makeatletter
\renewcommand{\@noticestring}{%
  Preprint. This is the submitted version, prior to peer review.
  Accepted at ECML-PKDD 2026 (Research Track).}
\makeatother

\usepackage[utf8]{inputenc} 
\usepackage[T1]{fontenc}    
\usepackage{hyperref}       
\usepackage{url}            
\usepackage{booktabs}       
\usepackage{amsfonts}       
\usepackage{nicefrac}       
\usepackage{microtype}      
\usepackage{xcolor}         

\usepackage{subfiles}
\usepackage{xcolor}
\usepackage{pifont}
\usepackage{array}
\usepackage{tikz}
\usepackage{algorithm}
\usepackage{amsmath}
\usepackage{algpseudocode}
\usepackage{makecell}
\usepackage{colortbl}
\usepackage{amsfonts}
\usepackage{multirow}
\usepackage{wrapfig}
\usepackage{enumitem}

\definecolor{tickgreen}{HTML}{2E8B57}
\definecolor{crossred}{HTML}{CC3333}
\definecolor{barcolor}{HTML}{4A90D9}
\definecolor{hlblue}{HTML}{D6EAF8}

\title{Spectral Transformation for Layer-wise Global Rank Discovery in Federated LoRA for Vision Transformers}

\author{%
  Hariharan Ramesh \quad\quad Jyotikrishna Dass \\
  Department of Electrical and Computer Engineering \\
  University of Arizona, Tucson, AZ 85719 \\
  \texttt{\{hariharanr, jdass\}@arizona.edu}
}

\begin{document}

\maketitle

\begin{abstract}
   Fine-tuning Vision Transformers (ViTs) with low-rank adapters (LoRA) promises better communication efficiency under federated setup, yet existing aggregation strategies face fundamental limitations. Independently averaging these LoRA factors is mathematically inconsistent, introducing cross-term aggregation error. In contrast, approaches that preserve heterogeneous client ranks by concatenating local adapters on the server substantially increase download cost and often require merging global LoRA updates into pretrained weights on the clients, leading to reinitialization lag, and unstable convergence. Other approaches further increase server-side overhead by reconstructing dense weight updates or training auxiliary models for refinement of aggregation error on the server. In this work, we propose \textbf{SpecTraL}, spectral transformation for layer-wise global rank discovery, that resolves these challenges within a unified design. SpecTraL stacks local LoRA modules from clients and performs orthonormal Householder Transformation of the stacked adapters directly in the low-rank latent space, eliminating the need for dense reconstruction of global model update and any additional refinement or auxillary training on the server. By leveraging the Spiked Covariance Model from Random Matrix Theory, SpecTraL analytically separates the global consensus signal from non-IID noise, discovering optimal layer-wise global ranks without manual hyperparameter tuning. To match local ranks and encourage stable fine-tuning for subsequent rounds of local updates on the clients, we introduce a padding-aware initialization framework that allows them to incorporate residual LoRA dimensions without re-merging those into pre-trained base model. Experiments on federated fine-tuning of ViT-B/16 and ViT-L/16 over DomainNet and NICO++ demonstrate improved accuracy–communication trade-offs, reduced server computation, and elimination of hyperparameter search for rank selection. Our code is publicly available at \texttt{\url{https://github.com/DASS-Lab-Group/SpecTraL}}.
 
\end{abstract}

\section{Introduction}\label{sec:intro}

Vision Transformers (ViTs)~\cite{dosovitskiy2021image} pre-trained on large-scale datasets~\cite{radford2021learning,oquab2024dinov2} have become the dominant paradigm for visual recognition, yet adapting them to downstream tasks through full fine-tuning remains computationally expensive. Parameter-efficient fine-tuning (PEFT) methods~\cite{houlsby2019parameter,hu2021lora} address this by updating only a small number of additional parameters while keeping the backbone frozen. Among them, Low-Rank Adaptation (LoRA)~\cite{hu2021lora} is widely adopted: it injects trainable low-rank matrices $\mathbf{B}\in\mathbb{R}^{m\times r}$ and $\mathbf{A}\in\mathbb{R}^{r\times n}$ into frozen Transformer layers, greatly reducing memory and compute requirements. When training data is distributed across institutions or edge devices, Federated Learning (FL)~\cite{mcmahan2017communication} enables collaborative fine-tuning by training locally on clients and aggregating updates at a central server. Combining LoRA with FL provides a communication-efficient solution for distributed fine-tuning, but introduces new challenges arising from \emph{data heterogeneity}~\cite{li2020federated,karimireddy2020scaffold,li2021model,acar2021federated} (differences in local data distributions) and \emph{computational heterogeneity}~\cite{cho2024heterogeneous,nguyen2022federated,diao2021heterofl} (differences in client resources). Recent work has begun integrating LoRA into FL frameworks~\cite{zhang2023federatedgpt,sun2024improving,he2024flora,bai2024federated,ramesh2025florist}, but effective aggregation of LoRA adapters remains challenging.

An effective federated LoRA aggregation mechanism must simultaneously satisfy five requirements:  
(1) \emph{exact aggregation}, recovering the true update $\mathbf{\Delta W^*}=\sum_k p_k\mathbf{B}_k\mathbf{A}_k$;  
(2) support for \emph{computational heterogeneity} via different client ranks;  
(3) \emph{communication efficiency} through compact global adapters;  
(4) \emph{tractable server computation}; and  
(5) \emph{spectral noise suppression}, since non-IID client drift introduces spurious directions in the aggregated spectrum.


\begin{figure}[t]
\centering
\includegraphics[width=\linewidth]{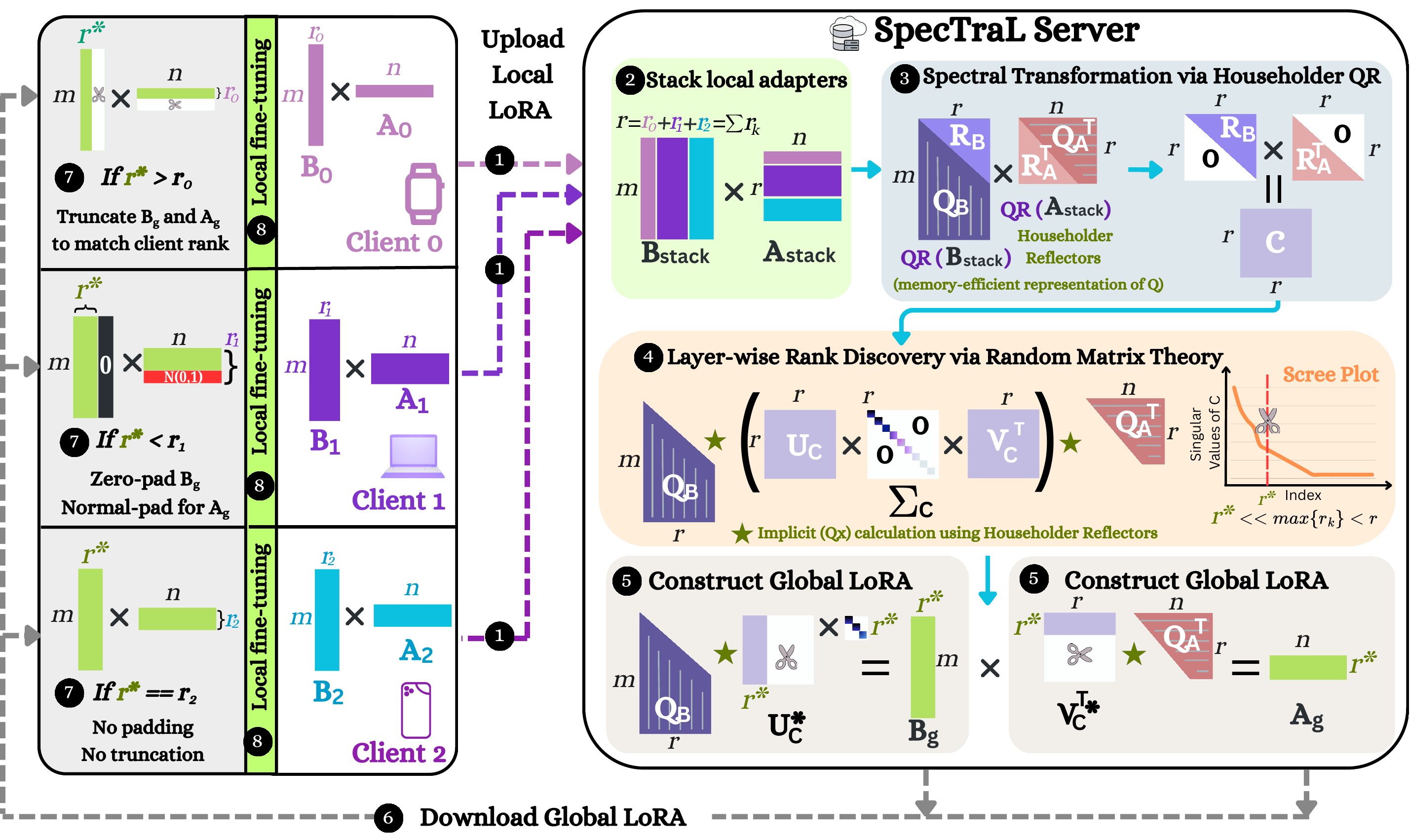}
\caption{Overview of SpecTraL: spectral transformation and layer-wise rank discovery for federated LoRA aggregation.}
\label{fig:overview}
\vspace{-8pt}
\end{figure}

Existing methods satisfy only subsets of these requirements. FedIT~\cite{zhang2023federatedgpt} averages $B$ and $A$ independently, producing $\mathbf{\bar{B}}\mathbf{\bar{A}}\neq\frac{1}{K}\sum_k\mathbf{B}_k\mathbf{A}_k$ and introducing cross-term noise. FFA-LoRA~\cite{sun2024improving} restores exact aggregation by freezing $A$, but halves the trainable parameter space and eliminates rank heterogeneity. FLoRA~\cite{he2024flora} preserves the exact update via stacking ($\mathbf{\Delta W^*}=\mathbf{B_{\mathrm{stack}}}\mathbf{A_{\mathrm{stack}}}$) and supports heterogeneous ranks, but broadcasts the full stacked adapters to all clients, inflating communication and requiring reinitialization of local adapters. LoRA-FAIR~\cite{bian2025lorafair} attempts to correct averaging bias by optimizing a residual $\Delta B$ at the server, increasing computation and abandoning rank heterogeneity while leaving signal and noise entangled in the spatial domain. FlexLoRA~\cite{bai2024federated} reconstructs the dense update $\mathbf{\Delta W}\in\mathbb{R}^{m\times n}$ and applies SVD to redistribute components according to client ranks, incurring prohibitive server cost and truncating singular values to match capacity rather than remove noise. FLoRIST~\cite{ramesh2025florist} improves efficiency by performing SVD in a compact stacked space and applying energy-based thresholding, revealing that the intrinsic dimensionality of aggregated updates is far smaller than the stacked rank. However, its empirically tuned threshold $\tau$ varies across models and datasets, leaving principled noise separation unresolved.

We propose \emph{Spec}tral \emph{Tra}nsformation for \emph{L}ayer-wise global rank discovery (\textbf{SpecTraL}), a federated LoRA aggregation framework that satisfies all five requirements. SpecTraL builds upon stacking-based aggregation, preserving exact updates and heterogeneous ranks. To recover the spectral structure of the aggregated update efficiently, it performs Householder QR decompositions of stacked factors ($\mathbf{B_{\mathrm{stack}}}=\mathbf{Q_B}\mathbf{R_B}$, $\mathbf{A_{\mathrm{stack}}}^T=\mathbf{Q_A}\mathbf{R_A}$) and computes a single SVD of the compact interaction matrix $\mathbf{C}=\mathbf{R_B}\mathbf{R_A}^T$, yielding the exact singular values of $\mathbf{\Delta W^*}$ at substantially lower cost than SVD on the stacked matrices.

SpecTraL then applies the ScreeNOT estimator~\cite{donoho2023screenot}, a random-matrix-theory-based method that identifies the MSE-optimal cutoff between signal and noise in the singular spectrum without manual tuning. The resulting threshold separates consensus directions shared across clients from incoherent components introduced by heterogeneous data. Because this procedure operates per layer and per weight matrix, SpecTraL naturally discovers layer-adaptive ranks reflecting the intrinsic dimensionality of each aggregated update. The resulting adapters are compact, improving communication efficiency as a consequence of principled denoising. When thresholding produces global adapters smaller than a client’s local rank, a padding-aware initialization strategy introduces exploratory directions in the orthogonal complement of the retained signal subspace, enabling stable local optimization.

\smallskip
\noindent\textbf{Contributions.} Our main contributions are:

\begin{enumerate}[leftmargin=*,itemsep=2pt,topsep=2pt]
\item \textbf{QR-accelerated spectral aggregation.}
We introduce an efficient aggregation pipeline that replaces SVD on stacked adapters with Householder QR decompositions followed by a compact SVD.

\item \textbf{Principled spectral denoising.}
We apply the ScreeNOT estimator to federated LoRA aggregation, replacing manually tuned energy thresholds with statistically grounded noise separation.

\item \textbf{Layer-adaptive rank discovery.}
We demonstrate that ViT layers exhibit distinct intrinsic dimensionalities and automatically determine per-layer ranks from the aggregated spectrum.

\item \textbf{Adapter re-initialization strategies:} We conduct a systematic study of five initialization methods for mapping compressed global updates back to client capacities. Our findings identify \textbf{Gaussian} as superior for maintaining exploratory capacity efficiently, while demonstrating that the standard \textbf{Zero-padding} approach significantly bottlenecks convergence.

\item \textbf{Extensive federated evaluation.}
We evaluate SpecTraL on ViT-B/16 and ViT-L/16 across DomainNet and NICO++ under realistic non-IID heterogeneity, outperforming FedIT, FFA-LoRA, FLoRA, FlexLoRA, FLoRIST, and LoRA-FAIR.
\end{enumerate}
\section{Preliminaries and Motivation}\label{sec:prelim}

\subsection{Federated Fine-Tuning with LoRA}\label{ssec:setup}

\noindent\textbf{LoRA.}
Given a pre-trained weight matrix $\mathbf{W_0} \in \mathbb{R}^{m \times n}$, Low-Rank Adaptation~\cite{hu2021lora} freezes $\mathbf{W}_0$ and introduces a trainable update $\mathbf{\Delta W}=\mathbf{B}\mathbf{A}$, where $\mathbf{B}\in\mathbb{R}^{m\times r}$, $\mathbf{A}\in\mathbb{R}^{r\times n}$, and $r\ll\min(m,n)$. The forward pass becomes $\mathbf{y}=(\mathbf{W_0}+\mathbf{B}\mathbf{A})\mathbf{x}$, reducing trainable parameters from $mn$ to $r(m+n)$. By convention, $\mathbf{B}$ is initialized to zero and $\mathbf{A}$ is drawn from a Gaussian distribution.

\smallskip
\noindent\textbf{Federated LoRA.}
In a federation of $K$ clients, each client $k$ holds a dataset $\mathcal{D}_k$ and trains local adapters $(\mathbf{B}_k,\mathbf{A}_k)$ with rank $r_k$. After local training, clients upload their adapters to a server, which must produce a global update representing the collective learning. The ideal aggregated update is

\begin{equation}
\mathbf{\Delta W^*}=\sum_{k=1}^{K} p_k\,\mathbf{B}_k\mathbf{A}_k,
\qquad
p_k=\frac{|\mathcal{D}_k|}{\sum_{j=1}^{K}|\mathcal{D}_j|}
\end{equation}

The server then broadcasts global adapters for the next round. The challenge is to compute $\mathbf{\Delta W^*}$ accurately, represent it compactly, and distribute it efficiently under heterogeneous client ranks.

\subsection{Federated LoRA Aggregation Methods}\label{ssec:aggregation_landscape}

\smallskip
\noindent\textbf{(i) Independent Averaging (FedIT~\cite{zhang2023federatedgpt}).}
FedIT averages LoRA factors independently: $\mathbf{\bar B}=\sum_k p_k\mathbf B_k$ and $\mathbf{\bar A}=\sum_k p_k\mathbf A_k$. The resulting update
$\bar{\mathbf B}\bar{\mathbf A}=\sum_k p_k^2\mathbf B_k\mathbf A_k+\sum_{i\neq j}p_ip_j\mathbf B_i\mathbf A_j$
contains cross-term noise that introduces spurious directions accumulating across rounds. FedIT also assumes uniform ranks; heterogeneous ranks require zero-padding~\cite{cho2024heterogeneous}, increasing communication cost.

\smallskip
\noindent\textbf{(ii) Freezing (FFA-LoRA~\cite{sun2024improving}).}
FFA-LoRA removes cross-terms by fixing $A$ to its initialization $A_{\mathrm{init}}$, so averaging $B$ alone exactly recovers $\mathbf{\Delta W^*}$. However, freezing $A$ halves the trainable parameter space and prevents heterogeneous ranks.

\smallskip
\noindent\textbf{(iii) Stacking (FLoRA~\cite{he2024flora}).}
FLoRA avoids cross-term noise by stacking adapters:
$
\mathbf{B_{\mathrm{stack}}}
=
[\mathbf{B}_1|\cdots|\mathbf{B}_K],\quad
\mathbf{A_{\mathrm{stack}}}
=
\begin{bmatrix}
p_1\mathbf A_1\\
\vdots\\
p_K\mathbf A_K
\end{bmatrix}
$
with $r=\sum_k r_k$. This guarantees $\mathbf{B_{\mathrm{stack}}}\mathbf{A_{\mathrm{stack}}}=\mathbf{\Delta W^*}$, enabling exact aggregation with heterogeneous rank support. However, the full stacked adapters are broadcast to all clients, so download cost grows linearly with $K$. Because no spectral mixing occurs in this representation, clients cannot truncate adapters to their local rank and must merge them into base weights, $\mathbf{W_0}\leftarrow\mathbf{W_0}+\mathbf{B_{\mathrm{stack}}}\mathbf{A_{\mathrm{stack}}}$, before reinitializing local adapters.

\smallskip
\noindent\textbf{(iv) Reconstruction + SVD (FlexLoRA~\cite{bai2024federated}).}
FlexLoRA reconstructs the dense update $\mathbf{\Delta W}\in\mathbb{R}^{m\times n}$ and applies SVD to redistribute truncated adapters to each client. This incurs $O(\min(m,n)\cdot mn)$ cost and requires storing the full matrix. More importantly, truncation matches client capacity ($r_k$) rather than the intrinsic dimensionality of the update.

\smallskip
\noindent\textbf{(v) Efficient SVD + Energy Thresholding (FLoRIST~\cite{ramesh2025florist}).}
FLoRIST retains stacking but recovers singular values in a compact space. It then selects the smallest rank $r^*$ satisfying $
\frac{\sum_{i=1}^{r^*}\sigma_i^2}
{\sum_{i=1}^{r}\sigma_i^2}
\ge\tau
$
producing compact global adapters. While this reveals that the intrinsic dimensionality is far smaller than the stacked rank, $\tau$ must be manually tuned and varies widely across models, datasets, and heterogeneity levels.

\smallskip
\noindent\textbf{(vi) Residual Correction (LoRA-FAIR~\cite{bian2025lorafair}).}
LoRA-FAIR stays within the averaging framework but learns a residual correction

\begin{equation}
\min_{\mathbf{\Delta B}}
\mathcal{S}\!\left(
\mathbf{\Delta W^*},
(\mathbf{\bar B}+\mathbf{\Delta B})\mathbf{\bar A}
\right)
+
\vartheta\|\mathbf{\Delta B}\|
\end{equation}

where $\mathcal{S}(\cdot,\cdot)$ denotes cosine similarity. While this reduces aggregation bias and initialization lag, it introduces iterative server optimization, increases computation, and abandons heterogeneous rank support.

\smallskip
\noindent\textbf{The missing piece.}
The progression from averaging (FedIT) to stacking (FLoRA) to spectral compression (FLoRIST) has steadily improved federated LoRA aggregation, yet one key question remains: \emph{given the singular value spectrum of the aggregated update, where does the task-relevant signal end and the noise begin?} FLoRIST shows that aggressive rank reduction can improve performance, suggesting that trailing singular values correspond to harmful directions, but relies on manually tuned thresholds and provides no principled way to identify the signal–noise boundary. Moreover, this spectral structure varies across layers and weight matrices, while existing methods enforce a fixed rank or global threshold. SpecTraL is designed to close these gaps.

\section{Proposed Method: SpecTraL}\label{sec:method}
\vspace{-8pt}
We present SpecTraL in three parts: a motivating empirical observation (\S\ref{ssec:motivation}), the full methodology (\S\ref{ssec:methodology}), and a complexity analysis (\S\ref{ssec:complexity}).
\vspace{-8pt}
\subsection{Motivating Observation}\label{ssec:motivation}

\begin{wrapfigure}{r}{0.65\textwidth}
    \vspace{-12pt}
    \centering
    \includegraphics[width=\linewidth]{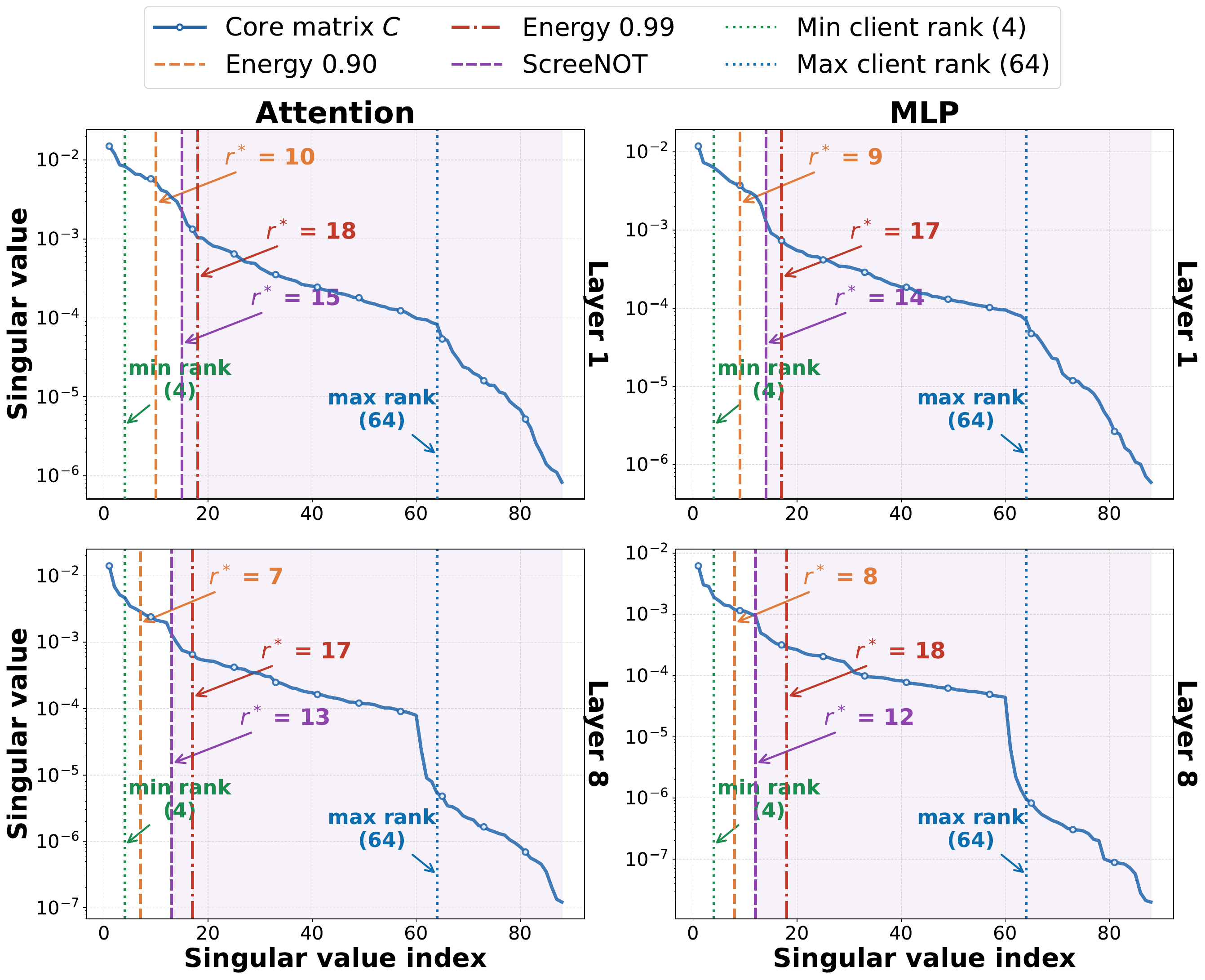}
    \caption{Singular value spectrum of the core matrix $\mathbf{C}$ (which shares the singular values of $\mathbf{\Delta W^*}$; see \S\ref{ssec:methodology}) for two representative layers of ViT-L/16 on Nico++ under heterogeneous client ranks. Columns correspond to attention and MLP weight matrices; rows to layers~1 and~8. Vertical lines mark the ranks selected by energy thresholds $\tau = 0.90$ (orange, dashed) and $\tau = 0.99$ (red, dash-dot), ScreeNOT (purple, dashed), the minimum client rank (green, 4), and the maximum client rank (blue, 64). ScreeNOT consistently identifies the elbow between signal and noise without manual tuning.}
    \label{fig:scree_grid}
    \vspace{-8pt}
\end{wrapfigure}

To understand why principled rank selection is necessary, we examine the singular value spectrum of the aggregated update $\mathbf{\Delta W^*}$ across different layers and weight types. Figure~\ref{fig:scree_grid} plots the singular values of the core matrix $\mathbf{C}$ for attention and MLP weight matrices at two layers of ViT-L/16 on Nico++. Three observations motivate our design.


\smallskip
\noindent\textbf{Different layers and weight types exhibit significantly different intrinsic dimensionalities.} 
As shown in Figure~\ref{fig:scree_grid}, the rank required to capture a specific amount of signal variance fluctuates substantially across the model. 
For instance, at an energy threshold of $\tau = 0.90$, the attention matrix in Layer~1 requires a rank of $r^* = 10$, whereas the same component in Layer~8 requires only $r^* = 7$, a 30\% reduction in necessary parameters for the same relative information. 
Across the entire grid for $\tau = 0.90$, the optimal rank varies from $r^* = 7$ to $18$. 
Even the automated ScreeNOT selector identifies non-uniform dimensionalities (e.g., $r^* = 15$ for Layer~1 Attention vs. $r^* = 12$ for Layer~8 MLP). 
These fluctuations confirm that a fixed rank applied uniformly to all layers, the default strategy for nearly all existing methods, inevitably leads to a sub-optimal trade-off: it either wastes communication bandwidth on over-parameterized layers or discards critical signal in layers with higher intrinsic dimensionality.

\smallskip
\noindent\textbf{Fixed energy thresholds are unreliable.} At $\tau = 0.90$, the attention matrix in Layer~8 retains only $r^* = 7$ components, cutting well into the signal region and losing information. At $\tau = 0.99$, the same layer retains $r^* = 17$, preserving noise components that degrade accuracy. No single $\tau$ works well across all layers simultaneously: FLoRIST's reported optimal threshold varies from 0.80 to 0.99 depending on the model, dataset, and heterogeneity level~\cite{ramesh2025florist}, requiring exhaustive search for each configuration.

\noindent\textbf{ScreeNOT identifies the signal--noise boundary automatically.} Across all four panels, ScreeNOT places its threshold at the elbow where dominant singular values transition into a slowly decaying tail. This elbow corresponds to the boundary between directions of inter-client consensus (the task-relevant signal) and incoherent, client-specific directions amplified by non-IID data heterogeneity. The effectiveness of this automatic separation is validated empirically: SpecTraL with ScreeNOT achieves the highest average accuracy across all model--dataset combinations in our experiments (Table~\ref{tab:main_homo}).

\subsection{Methodology}\label{ssec:methodology}

SpecTraL operates in four steps per communication round: (1)~spectral transformation of heterogeneous local LoRA updates via Householder QR, (2)~discovery of layer-wise global ranks via ScreeNOT, (3)~efficient construction and broadcast of compact global adapters, and (4)~residual initialization of local LoRA to match client ranks for subsequent rounds.

\medskip
\noindent\underline{\textbf{Step 1: Spectral Transformation via Householder QR.}}

\smallskip
Like FLoRA~\cite{he2024flora} and FLoRIST~\cite{ramesh2025florist}, SpecTraL begins with noise-free stacking-based aggregation, producing $\mathbf{B}_{\mathrm{stack}} \in \mathbb{R}^{m \times r}$ and $\mathbf{A}_{\mathrm{stack}} \in \mathbb{R}^{r \times n}$ with $r = \sum_k r_k$, where $r \ll \max\{m, n\}$. Our goal is to recover the singular values of the true aggregated update $\mathbf{\Delta W^*} = \mathbf{B}_{\mathrm{stack}} \mathbf{A}_{\mathrm{stack}}$ without forming the full $m \times n$ matrix. FLoRIST achieves this by performing thin SVD on $\mathbf{B}_{\mathrm{stack}}$ and $\mathbf{A}_{\mathrm{stack}}$ independently. We observe that SVD is more expensive than necessary: QR decomposition suffices to extract orthonormal bases at lower cost.

We compute thin QR decompositions via Householder reflectors~\cite{doi:10.1137/1.9781421407944}:
\begin{equation}\label{eq:qr}
    \mathbf{B}_{\mathrm{stack}} = \mathbf{Q}_B \mathbf{R}_B, \qquad \mathbf{A}_{\mathrm{stack}}^T = \mathbf{Q}_A \mathbf{R}_A,
\end{equation}
where $\mathbf{Q}_B \in \mathbb{R}^{m \times r}$ and $\mathbf{Q}_A \in \mathbb{R}^{n \times r}$ have orthonormal columns, and $\mathbf{R}_B, \mathbf{R}_A \in \mathbb{R}^{r \times r}$ are upper triangular. The Householder variant stores the orthonormal factors implicitly as products of $r$ reflectors, never forming $\mathbf{Q}$ explicitly. Each thin QR costs $O(mr^2)$ and $O(nr^2)$ flops, respectively.

We then form the core matrix:
\begin{equation}\label{eq:core}
    \mathbf{C} = \mathbf{R}_B \, \mathbf{R}_A^T \in \mathbb{R}^{r \times r}.
\end{equation}
This $r \times r$ matrix captures all the spectral information of $\mathbf{\Delta W^*}$. To see why, substitute Eq.~\ref{eq:qr} into the aggregated update:
\begin{equation}\label{eq:equivalence}
    \mathbf{\Delta W^*} = \mathbf{Q}_B \mathbf{R}_B (\mathbf{Q}_A \mathbf{R}_A)^T = \mathbf{Q}_B \, \underbrace{(\mathbf{R}_B \mathbf{R}_A^T)}_{\mathbf{C}} \, \mathbf{Q}_A^T.
\end{equation}
Since $\mathbf{Q}_B$ and $\mathbf{Q}_A$ have orthonormal columns, the singular values of $\mathbf{\Delta W^*}$ are exactly the singular values of $\mathbf{C}$. We compute the SVD of this small matrix:
\begin{equation}\label{eq:svd_core}
    \mathbf{C} = \mathbf{U}_C \, \mathbf{\Sigma}_C \, \mathbf{V}_C^T,
\end{equation}
where $\mathbf{\Sigma}_C = \mathrm{diag}(\sigma_1, \ldots, \sigma_r)$ with $\sigma_1 \geq \cdots \geq \sigma_r \geq 0$. This SVD costs $O(r^3)$, independent of the model dimensions $m$ and $n$. The diagonal entries $\sigma_i$ are the exact singular values of the aggregated LoRA update $\mathbf{\Delta W^*}$, computed without ever forming $\mathbf{\Delta W^*}$ itself.

\medskip
\noindent\underline{\textbf{Step 2: Discovering Layer-wise Global LoRA Rank via ScreeNOT.}}

\smallskip
The singular values $\{\sigma_i\}_{i=1}^r$ of $\mathbf{C}$ reflect a superposition of two components: a low-rank signal matrix $\mathbf{\Delta W}_{\mathrm{signal}}$ encoding the directions of inter-client consensus (the task-relevant update), and a perturbation matrix $\mathbf{\Delta W}_{\mathrm{noise}}$ comprising incoherent, client-specific directions amplified by non-IID data heterogeneity. In random matrix theory, this structure is formalized as the \emph{spiked covariance model}~\cite{johnstone2001distribution}: the observed matrix $\mathbf{C}$ is modeled as a low-rank signal corrupted by noise whose singular values follow a compactly supported distribution (e.g., Marchenko--Pastur). The leading singular values of $\mathbf{C}$ that ``spike'' above this noise distribution correspond to the signal, while the bulk corresponds to noise.

The challenge is to identify the cutoff $\hat{\theta}$ that separates the two. In our federated setting, the noise in $\mathbf{C}$ is \emph{not} white: heterogeneous clients with different data distributions, training durations, and local ranks produce correlated perturbations whose covariance structure is unknown. ScreeNOT~\cite{donoho2023screenot} is designed precisely for this scenario. Given the observed singular values $\sigma_1 \geq \cdots \geq \sigma_r$ of $\mathbf{C}$, the matrix dimensions, and a loose upper bound $k$ on the signal rank, ScreeNOT returns the MSE-optimal hard threshold $\hat{\theta}_{\mathrm{SN}}$. The resulting rank:
\begin{equation}\label{eq:screenot_rank}
    r^* = \big|\{i : \sigma_i > \hat{\theta}_{\mathrm{SN}}\}\big|
\end{equation}
is the number of singular components to retain. ScreeNOT guarantees that the reconstructed signal $\mathbf{\hat{\Delta W}}_{\mathrm{signal}} = \sum_{i=1}^{r^*} \sigma_i \, \mathbf{u}_i \mathbf{v}_i^T$ achieves the lowest possible MSE among all hard-threshold estimators:
\begin{equation}\label{eq:screenot_opt}
    \big\|\mathbf{\hat{\Delta W}}_{\mathrm{signal}} - \mathbf{\Delta W}_{\mathrm{signal}}\big\|_F^2 = \min_{\theta} \Big\| \sum_{i:\sigma_i > \theta} \sigma_i \, \mathbf{u}_i \mathbf{v}_i^T - \mathbf{\Delta W}_{\mathrm{signal}} \Big\|_F^2,
\end{equation}
with probability tending to 1 as the matrix dimensions grow~\cite{donoho2023screenot}. The computational cost of ScreeNOT is $O(r \log r)$ (dominated by sorting), negligible compared to the QR and SVD steps.

Because ScreeNOT operates on the singular values of $\mathbf{C}$ independently at each layer $l$ and each weight matrix type ($\mathbf{W}_{\mathrm{attn}}$, $\mathbf{W}_{\mathrm{mlp}}$) at every communication round $t$, SpecTraL automatically produces a layer-specific rank $r_l^{*}(t)$ that reflects the intrinsic dimensionality of that layer's aggregated update at that round. No global rank hyperparameter is needed. This is the \emph{layer-wise global rank discovery} that gives SpecTraL its name.

\medskip
\noindent\underline{\textbf{Step 3: Efficient Construction and Broadcast of Global LoRA.}}

\smallskip
Having identified the optimal rank $r^*$, we construct compact global adapters \emph{without} explicitly forming the orthonormal matrices $\mathbf{Q}_B$ or $\mathbf{Q}_A$. Let $\tilde{\mathbf{U}} = (\mathbf{U}_C)_{:,1:r^*} \in \mathbb{R}^{r \times r^*}$ and $\tilde{\mathbf{V}} = (\mathbf{V}_C)_{:,1:r^*} \in \mathbb{R}^{r \times r^*}$ denote the leading $r^*$ columns of the SVD factors from Step~1. The global adapters are:
\begin{equation}\label{eq:global_adapters}
    \mathbf{B}_g = \mathcal{Q}_B(\tilde{\mathbf{U}}) \cdot (\mathbf{\Sigma}_C)_{1:r^*,\,1:r^*} \in \mathbb{R}^{m \times r^*}, \qquad \mathbf{A}_g = \big(\mathcal{Q}_A(\tilde{\mathbf{V}})\big)^T \in \mathbb{R}^{r^* \times n},
\end{equation}
where $\mathcal{Q}_B(\cdot)$ denotes the application of the stored Householder reflectors from $\mathbf{B}_{\mathrm{stack}}$'s QR factorization to a given matrix (i.e., computing $\mathbf{Q}_B \tilde{\mathbf{U}}$ implicitly), and $\mathcal{Q}_A(\cdot)$ is defined analogously for $\mathbf{A}_{\mathrm{stack}}^T$. In practice, this corresponds to a call to \texttt{DORMQR}, which applies the reflectors to the $r \times r^*$ matrix $\tilde{\mathbf{U}}$ at a cost of $O(mr \cdot r^*)$, significantly cheaper than forming the full $m \times r$ matrix $\mathbf{Q}_B$ at $O(mr^2)$ when $r^* \ll r$.

The resulting adapters satisfy $\mathbf{B}_g \mathbf{A}_g \approx \mathbf{\Delta W^*}$, with the approximation error consisting precisely of the thresholded noise components. The server broadcasts the compact pair $(\mathbf{B}_g, \mathbf{A}_g)$ to all clients. Because $r^*$ reflects the true signal complexity rather than an arbitrary hyperparameter, the broadcast cost is naturally minimized: communication efficiency improves as a direct consequence of principled denoising.

\medskip
\noindent\underline{\textbf{Step 4: Residual Initialization of Local LoRA.}}

\smallskip
After receiving the global adapters $(\mathbf{B}_g, \mathbf{A}_g)$ of rank $r^*$, each client $k$ must resume local training at its full rank $r_k$. Since typically $r^* \ll r_k$ after aggressive spectral denoising, the client pads the received adapters with $r_k - r^*$ additional dimensions. We initialize the padded rows of $\mathbf{A}$ with random Gaussian vectors while zero-padding $\mathbf{B}$:
\begin{equation}\label{eq:gaussian_init}
    \mathbf{B}_k = [\mathbf{B}_g \mid \mathbf{0}_{m \times (r_k - r^*)}], \qquad \mathbf{A}_k = \begin{bmatrix} \mathbf{A}_g \\ \mathbf{A}_{\mathrm{new}} \end{bmatrix}, \quad \mathbf{A}_{\mathrm{new}} \sim \mathcal{N}(0, \sigma_A^2),
\end{equation}
where $\sigma_A$ is set to match the standard deviation of the existing rows of $\mathbf{A}_g$. This design has two important properties. First, the initial forward-pass contribution $\mathbf{\Delta W} = \mathbf{B}_k \mathbf{A}_k = \mathbf{B}_g \mathbf{A}_g$ remains unchanged because the zero columns in $\mathbf{B}_k$ annihilate the Gaussian rows in $\mathbf{A}_k$, preserving the denoised global signal exactly. Second, unlike zero-padding (which provides no gradient signal to the new dimensions), the Gaussian rows in $\mathbf{A}_k$ provide diverse exploratory directions that receive immediate gradient updates through $\mathbf{B}_k$, enabling the client to adapt locally beyond the global signal subspace from the first step of training. We refer to this as \emph{residual initialization} because the padded dimensions represent the residual capacity available for client-specific exploration on top of the globally shared signal. Alternative initialization strategies (zero-padding, orthogonal complement, pretrained-SVD) are compared in the ablation study (\S\ref{ssec:ablations}).

\subsection{Complexity Analysis}\label{ssec:complexity}

The full SpecTraL pipeline for one communication round is summarized in Algorithm~\ref{alg:spectral}. We analyze the per-round server-side cost below.

\smallskip
\noindent\textbf{Step 1 (QR + core matrix + SVD).}
Two thin Householder QR factorizations (without forming $\mathbf{Q}$) cost $2(m+n)r^2 - \tfrac{4}{3}r^3$ flops. Forming $\mathbf{C} = \mathbf{R}_B \mathbf{R}_A^T$ costs $O(r^3)$. The SVD of the $r \times r$ matrix $\mathbf{C}$ costs $O(r^3)$.

\smallskip
\noindent\textbf{Step 2 (ScreeNOT).}
Computing the threshold costs $O(r \log r)$, dominated by sorting the singular values.

\smallskip
\noindent\textbf{Step 3 (Reconstruction).}
Applying stored Householder reflectors to the $r \times r^*$ matrices $\tilde{\mathbf{U}}$ and $\tilde{\mathbf{V}}$ costs $O((m+n) r \cdot r^*)$.

\smallskip
\noindent\textbf{Total server cost per layer.}
Summing the dominant terms:
\begin{equation}\label{eq:spectral_cost}
    \underbrace{2(m+n)r^2 - \tfrac{4}{3}r^3}_{\text{QR factorization}} + \underbrace{O(r^3)}_{\text{core SVD}} + \underbrace{O\big((m+n) r \cdot r^*\big)}_{\text{reconstruction}}.
\end{equation}
For comparison, FLoRIST's server cost is dominated by two thin SVDs on the stacked factors at $14(m+n)r^2 + 16r^3$ flops plus $O(r^3)$ for the interaction matrix SVD. The QR factorization step alone is over $7\times$ cheaper in the leading $(m+n)r^2$ term, and the reconstruction cost is further reduced when $r^* \ll r$ (as is typical after ScreeNOT thresholding). FlexLoRA's cost of $O(\min(m,n) \cdot mn)$ for full dense SVD is orders of magnitude higher. Since $r = \sum_k r_k$ is typically $\ll \min(m,n)$ for LoRA configurations and $r^* \ll r$ after thresholding, SpecTraL's server computation remains tractable even for large Vision Transformers.

\begin{algorithm}[t]
\caption{SpecTraL: One Communication Round}\label{alg:spectral}
\begin{algorithmic}[1]
\Require Clients $\{1, \ldots, K\}$ with adapters $\{(\mathbf{B}_k, \mathbf{A}_k)\}$, ScreeNOT upper bound $k$
\Ensure Global adapters $(\mathbf{B}_g, \mathbf{A}_g)$ for each adapted layer
\For{each adapted layer $l$}
    \State \textbf{Stack:} $\mathbf{B}_{\mathrm{stack}} \leftarrow [\mathbf{B}_1 \mid \cdots \mid \mathbf{B}_K]$, \; $\mathbf{A}_{\mathrm{stack}} \leftarrow [p_1 \mathbf{A}_1; \ldots; p_K \mathbf{A}_K]$
    \State \textbf{QR:} $\mathbf{B}_{\mathrm{stack}} \xrightarrow{\text{Householder}} \mathbf{R}_B$, \; $\mathbf{A}_{\mathrm{stack}}^T \xrightarrow{\text{Householder}} \mathbf{R}_A$ \hfill \textit{(reflectors stored implicitly)}
    \State \textbf{Core:} $\mathbf{C} \leftarrow \mathbf{R}_B \mathbf{R}_A^T$
    \State \textbf{SVD:} $\mathbf{C} = \mathbf{U}_C \, \mathbf{\Sigma}_C \, \mathbf{V}_C^T$, \; singular values $\sigma_1 \geq \cdots \geq \sigma_r$
    \State \textbf{ScreeNOT:} $\hat{\theta} \leftarrow \mathrm{ScreeNOT}(\sigma_1, \ldots, \sigma_r;\, r,\, k)$, \; $r^* \leftarrow |\{i : \sigma_i > \hat{\theta}\}|$
    \State \textbf{Reconstruct:} $\mathbf{B}_g \leftarrow \mathcal{Q}_B\!\big((\mathbf{U}_C)_{:,1:r^*}\big) \cdot (\mathbf{\Sigma}_C)_{1:r^*,1:r^*}$, \; $\mathbf{A}_g \leftarrow \big(\mathcal{Q}_A\!\big((\mathbf{V}_C)_{:,1:r^*}\big)\big)^T$ \hfill \textit{(apply reflectors)}
\EndFor
\State \textbf{Broadcast} $(\mathbf{B}_g, \mathbf{A}_g)$ per layer to all clients
\For{each client $k$}
    \State \textbf{Pad:} $\mathbf{B}_k \leftarrow [\mathbf{B}_g \mid \mathbf{0}]$, \; $\mathbf{A}_k \leftarrow [\mathbf{A}_g;\, \mathbf{A}_{\mathrm{new}}]$ with $\mathbf{A}_{\mathrm{new}} \sim \mathcal{N}(0, \sigma_A^2)$ \hfill \textit{(Gaussian init)}
    \State \textbf{Train:} $(\mathbf{B}_k, \mathbf{A}_k) \leftarrow \mathrm{LocalUpdate}(\mathbf{W}_0, \mathbf{B}_k, \mathbf{A}_k, \mathcal{D}_k, E)$
\EndFor
\end{algorithmic}
\end{algorithm}


\section{Experiments}\label{sec:experiments}

We evaluate SpecTraL on federated fine-tuning of Vision Transformers across two large-scale benchmarks under realistic non-IID heterogeneity. We first describe the experimental setup (\S\ref{ssec:setup}), present the main performance comparison (\S\ref{ssec:main_results}), report ablation studies on initialization strategies (\S\ref{ssec:ablations}) and evaluate SpecTraL under computational heterogeneity (\S\ref{ssec:hetero}).

\subsection{Experimental Setup}\label{ssec:setup}

Table~\ref{tab:main_homo} reports per-domain accuracy on DomainNet and NICO++ with ViT-B/16 and ViT-L/16.

\begin{table}[t!]
\centering
\caption{Per-domain accuracy (\%) under feature and label non-IID with client rank $r\!=\!32$ (100 clients, 75 rounds). Best in \textbf{bold}, second-best \underline{underlined}. SpecTraL uses ScreeNOT + Gaussian initialization.}
\label{tab:main_homo}
\setlength{\tabcolsep}{3.5pt}
\renewcommand{\arraystretch}{1.15}
\small
\resizebox{\textwidth}{!}{%
\begin{tabular}{cl l cccccc c}
\toprule
\textbf{Model} & \textbf{Dataset} & \textbf{Method} & \multicolumn{6}{c}{\textbf{Per-Domain Accuracy (\%)}} & \cellcolor{black!8}\textbf{Avg.} \\
\midrule
& & & Clipart & Infograph & Painting & Quickdraw & Real & Sketch & \\
\cmidrule(lr){4-9}
\multirow{14}{*}{\rotatebox{90}{\textbf{ViT-B/16}}}
& \multirow{7}{*}{\rotatebox{90}{\textbf{DomainNet}}}
& FFA-LoRA     & 70.18 & 39.41 & 74.50 & 40.28 & 87.85 & 60.59 & \cellcolor{black!8}62.13 \\
&& FedIT        & 82.01 & 51.91 & \underline{79.88} & 72.64 & \textbf{90.86} & 77.34 & \cellcolor{black!8}75.77 \\
&& FLoRA        & 73.02 & 40.50 & 75.58 & 53.45 & 88.69 & 66.73 & \cellcolor{black!8}66.33 \\
&& FlexLoRA     & 81.98 & 51.64 & 79.60 & \underline{74.10} & 90.63 & \textbf{78.27} & \cellcolor{black!8}\underline{76.04} \\
&& LoRA-FAIR    & 53.02 & 21.94 & 47.96 & 55.61 & 68.62 & 42.09 & \cellcolor{black!8}48.21 \\
&& FLoRIST      & 81.80 & 50.20 & 79.10 & 72.69 & 90.35 & 77.34 & \cellcolor{black!8}75.25 \\
\cmidrule(lr){3-10}
&& \cellcolor{blue!8}\textbf{SpecTraL} & \cellcolor{blue!8}\textbf{82.29} & \cellcolor{blue!8}\textbf{51.94} & \cellcolor{blue!8}79.74 & \cellcolor{blue!8}\textbf{74.14} & \cellcolor{blue!8}\underline{90.72} & \cellcolor{blue!8}\underline{77.61} & \cellcolor{black!8}\textbf{76.21} \\
\cmidrule(lr){2-10}
& & & Autumn & Dim & Grass & Outdoor & Rock & Water & \\
\cmidrule(lr){4-9}
& \multirow{7}{*}{\rotatebox{90}{\textbf{NICO++}}}
& FFA-LoRA     & 89.33 & 83.46 & 91.26 & 88.04 & 87.52 & 87.34 & \cellcolor{black!8}87.83 \\
&& FedIT        & 90.65 & 88.39 & 93.16 & 89.99 & \underline{89.78} & \underline{90.60} & \cellcolor{black!8}90.43 \\
&& FLoRA        & 89.49 & 83.94 & 91.39 & 88.31 & 88.34 & 87.50 & \cellcolor{black!8}88.16 \\
&& FlexLoRA     & \underline{91.03} & \textbf{88.91} & \underline{93.67} & \underline{90.71} & 89.56 & 90.60 & \cellcolor{black!8}\underline{90.75} \\
&& LoRA-FAIR    & 74.71 & 69.61 & 79.56 & 77.53 & 74.64 & 76.07 & \cellcolor{black!8}75.35 \\
&& FLoRIST      & 91.14 & \underline{88.83} & \textbf{93.77} & 90.44 & 89.83 & 89.76 & \cellcolor{black!8}90.63 \\
\cmidrule(lr){3-10}
&& \cellcolor{blue!8}\textbf{SpecTraL} & \cellcolor{blue!8}\textbf{91.58} & \cellcolor{blue!8}88.19 & \cellcolor{blue!8}93.65 & \cellcolor{blue!8}\textbf{90.86} & \cellcolor{blue!8}\textbf{89.92} & \cellcolor{blue!8}\textbf{90.96} & \cellcolor{black!8}\textbf{90.86} \\
\midrule
& & & Clipart & Infograph & Painting & Quickdraw & Real & Sketch & \\
\cmidrule(lr){4-9}
\multirow{14}{*}{\rotatebox{90}{\textbf{ViT-L/16}}}
& \multirow{7}{*}{\rotatebox{90}{\textbf{DomainNet}}}
& FFA-LoRA     & 77.75 & 45.18 & 79.82 & 55.87 & 88.86 & 71.45 & \cellcolor{black!8}69.82 \\
&& FedIT        & 84.23 & 58.06 & \underline{84.30} & 76.35 & 91.11 & 81.50 & \cellcolor{black!8}79.26 \\
&& FLoRA        & 77.96 & 44.74 & 80.11 & 59.95 & 89.27 & 71.68 & \cellcolor{black!8}70.62 \\
&& FlexLoRA     & \underline{84.88} & \underline{58.32} & 84.25 & \underline{76.54} & \underline{91.31} & \underline{81.58} & \cellcolor{black!8}\underline{79.48} \\
&& LoRA-FAIR & 84.41 & 57.88 & 84.22 & 76.16 & 91.11 & 81.37 & \cellcolor{black!8}79.19 \\
&& FLoRIST      & 84.60 & 57.09 & 84.25 & 75.51 & 91.11 & 80.92 & \cellcolor{black!8}78.91 \\
\cmidrule(lr){3-10}
&& \cellcolor{blue!8}\textbf{SpecTraL} & \cellcolor{blue!8}\textbf{85.07} & \cellcolor{blue!8}\textbf{58.97} & \cellcolor{blue!8}\textbf{84.36} & \cellcolor{blue!8}\textbf{76.64} & \cellcolor{blue!8}\textbf{91.32} & \cellcolor{blue!8}\textbf{81.74} & \cellcolor{black!8}\textbf{79.68} \\
\cmidrule(lr){2-10}
& & & Autumn & Dim & Grass & Outdoor & Rock & Water & \\
\cmidrule(lr){4-9}
& \multirow{7}{*}{\rotatebox{90}{\textbf{NICO++}}}
& FFA-LoRA     & 91.20 & 87.79 & 92.62 & 90.53 & 89.60 & 89.22 & \cellcolor{black!8}90.16 \\
&& FedIT        & 93.51 & 90.75 & 94.56 & 91.82 & 92.27 & 91.69 & \cellcolor{black!8}92.43 \\
&& FLoRA        & 91.42 & 87.87 & 92.69 & 90.44 & 89.24 & 88.88 & \cellcolor{black!8}90.09 \\
&& FlexLoRA     & \textbf{94.00} & \textbf{92.51} & \textbf{95.17} & \underline{92.48} & \underline{92.36} & \underline{92.58} & \cellcolor{black!8}\underline{93.18} \\
&& LoRA-FAIR    & 90.87 & 85.94 & 93.26 & 88.94 & 88.79 & 88.96 & \cellcolor{black!8}89.46 \\
&& FLoRIST      & \underline{93.73} & 92.03 & 94.44 & 92.24 & 92.45 & 92.11 & \cellcolor{black!8}92.83 \\
\cmidrule(lr){3-10}
&& \cellcolor{blue!8}\textbf{SpecTraL} & \cellcolor{blue!8}93.73 & \cellcolor{blue!8}\underline{92.39} & \cellcolor{blue!8}\underline{95.13} & \cellcolor{blue!8}\textbf{92.86} & \cellcolor{blue!8}\textbf{92.95} & \cellcolor{blue!8}\textbf{93.02} & \cellcolor{black!8}\textbf{93.35} \\
\bottomrule
\end{tabular}%
}
\end{table}
A~~

\noindent\textbf{Models.}
We use two Vision Transformer architectures: ViT-B/16 and ViT-L/16~\cite{dosovitskiy2021image}, both pre-trained on ImageNet-21k and loaded from the \texttt{timm} library~\cite{rw2019timm}. LoRA adapters are applied to all attention projection matrices ($W_q$, $W_k$, $W_v$, $W_o$) and MLP layers ($W_{\mathrm{fc1}}$, $W_{\mathrm{fc2}}$) in each Transformer block. All clients train LoRA adapters at rank $r = 32$.

\smallskip
\noindent\textbf{Datasets.}
We evaluate on two real-world image classification benchmarks used in prior federated LoRA studies~\cite{bian2025lorafair}:
\textbf{DomainNet}~\cite{peng2019moment}, a large-scale multi-domain dataset containing ${\sim}600$k images across 345 categories distributed over six visual domains (clipart, infograph, painting, quickdraw, real, sketch), where we use the first 100 categories following~\cite{bian2025lorafair}; and
\textbf{NICO++}~\cite{he2021towards}, an enhanced non-IID image classification dataset containing ${\sim}90$k images across 60 categories representing six visual styles (autumn, dim, grass, outdoor, rock, water).

\smallskip
\noindent\textbf{Federated configuration.}
Our setup consists of 100 clients, with 10 randomly sampled each round. We use a \emph{feature and label non-IID} partition following~\cite{bian2025lorafair}: clients are grouped by domain (or style), with approximately 16--17 clients per domain. Within each group, label distributions are further skewed using a Dirichlet distribution with concentration parameter $\alpha = 0.5$~\cite{li2022federated_noniid}. All experiments run for 75 communication rounds with 1 local epoch per round, using SGD with learning rate 0.01 and mini-batch size 128.

\smallskip
\begin{sloppypar}
\noindent\textbf{Baselines.} 
We compare against six methods that represent the major paradigms in federated LoRA aggregation: 
\textbf{FedIT}~\cite{zhang2023federatedgpt} (independent averaging of $B$ and $A$),
\textbf{FFA-LoRA}~\cite{sun2024improving} (freezes $A$, averages only $B$),
\textbf{FLoRA}~\cite{he2024flora} (stacking-based exact aggregation),
\textbf{FlexLoRA}~\cite{bai2024federated} (full dense reconstruction + SVD),
\textbf{LoRA-FAIR}~\cite{bian2025lorafair} (averaging with server-side residual correction, $\lambda = 0.01$), and
\textbf{FLoRIST}~\cite{ramesh2025florist} (efficient SVD on stacked adapters with energy threshold $\tau = 0.95$ and zero-padding initialization).
\end{sloppypar}

\smallskip
\noindent\textbf{SpecTraL configuration.}
All SpecTraL results use ScreeNOT thresholding with Gaussian initialization of padded $A$ rows, which we identify as the strongest combination in our ablation study (\S\ref{ssec:ablations}). ScreeNOT requires no hyperparameter tuning; it takes only the observed singular values and matrix dimensions as input.

\smallskip
\noindent\textbf{Evaluation.}
We report per-domain (or per-style) classification accuracy on held-out test sets, following the protocol of~\cite{bian2025lorafair}.

\subsection{Main Results}\label{ssec:main_results}

\begin{wrapfigure}{r}{0.45\textwidth}
    \centering
    \includegraphics[width=\linewidth]{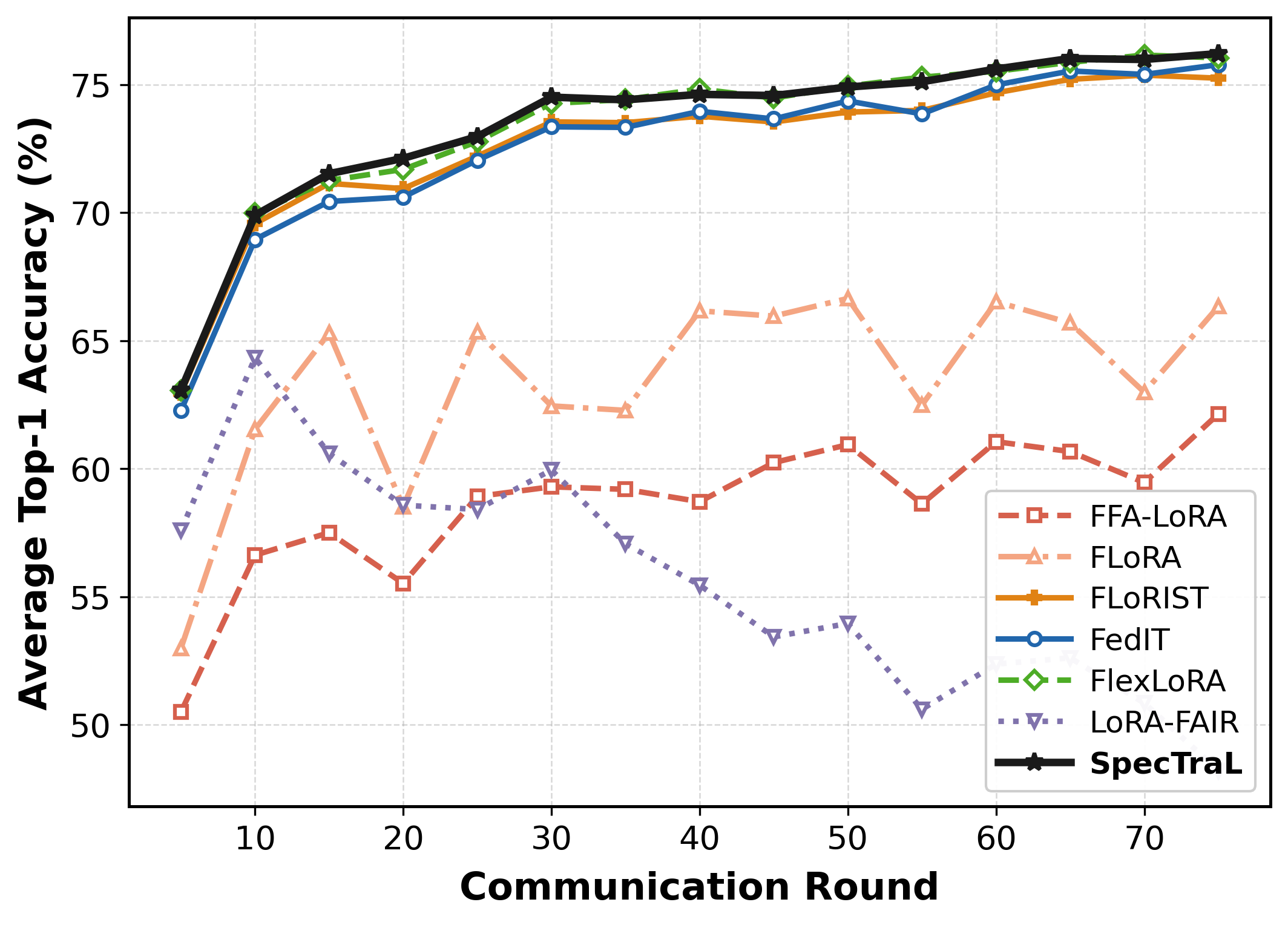}
    \caption{Training convergence on DomainNet with ViT-B/16 under homogeneous rank settings. SpecTraL consistently achieves higher accuracy throughout training and converges faster than existing federated LoRA aggregation methods.}
    \label{fig:domainnet_convergence}
    \vspace{-14pt}
\end{wrapfigure}

\noindent\textbf{SpecTraL consistently achieves the best average accuracy.}
Across all four model–dataset combinations, SpecTraL outperforms prior methods and achieves the highest average accuracy while maintaining strong performance across nearly every individual domain. These gains demonstrate that spectral denoising improves aggregation quality rather than favoring a subset of domains. As shown in Figure~\ref{fig:domainnet_convergence}, SpecTraL also maintains the highest accuracy throughout training, indicating faster and more stable convergence compared to existing aggregation methods.

\smallskip
\noindent\textbf{Principled thresholding improves over empirical energy thresholds.}
The comparison with FLoRIST isolates the effect of rank selection, as both methods use stacking-based aggregation. FLoRIST applies a fixed energy threshold ($\tau=0.95$) selected via binary search, whereas SpecTraL uses ScreeNOT to automatically determine the signal–noise boundary from the singular value spectrum. SpecTraL consistently outperforms FLoRIST across all settings without any threshold tuning, indicating that random-matrix-theory-based thresholding separates signal from noise more effectively than a fixed energy criterion.
\vspace{-14pt}

\smallskip
\noindent\textbf{Benefits increase with model scale.}
The performance gap between SpecTraL and FLoRIST widens on ViT-L/16 compared to ViT-B/16. Larger models produce richer singular value spectra across more adapted layers, making principled spectral denoising increasingly important.

\smallskip
\noindent\textbf{Averaging-based methods degrade under large heterogeneous federations.}
LoRA-FAIR collapses on ViT-B/16 DomainNet and remains below SpecTraL even on ViT-L/16, suggesting that server-side residual optimization cannot fully correct aggregation bias at scale. FFA-LoRA also underperforms across settings due to freezing half of the trainable parameters.

\smallskip
\noindent\textbf{Stacking alone is insufficient.}
While FLoRA avoids cross-term noise through stacking, its lack of spectral processing leads to weaker performance. Because the stacked representation lacks spectral mixing, clients must reinitialize local adapters each round, slowing convergence. SpecTraL instead broadcasts compact spectrally denoised adapters that clients directly use for initialization.

\subsection{Ablation Study: Initialization Strategies}\label{ssec:ablations}

We ablate the client-side initialization strategy for padded adapter dimensions using ScreeNOT~\cite{donoho2023screenot} for rank discovery. After thresholding, the global rank $r*$ is often smaller than the client rank $r_k$. While columns of $B$ are zero-padded, we investigate five strategies for the $r_k - r*$ additional rows of $A$:
(i)~\textbf{Zero-padding}: provides no exploratory directions; 
(ii)~\textbf{Gaussian}: $\mathcal{N}(0, \sigma_A^2)$ noise to provide random exploration; 
(iii)~\textbf{Orthogonal complement}: projects noise onto the orthogonal complement of $A_g$ for non-redundant directions; 
(iv)~\textbf{Pretrained-SVD}: samples from the trailing right singular vectors of $W_0$ to leverage pretrained structure; and 
(v)~\textbf{Trained-$A$}: reuses the client's previous local $A$ for continuity.


\smallskip
\noindent\textbf{Analysis.} 
Table~\ref{tab:ablation_init} shows that \textbf{Pretrained-SVD} achieves the highest accuracy (91.08\%), likely by leveraging task-relevant directions from the foundation model. However, \textbf{Gaussian} initialization (90.86\%) is a close second and offers a more practical default as it requires no access to $W_0$ or stale local state and no additional compute cost. Conversely, \textbf{Zero-padding} is consistently the weakest (89.86\%), confirming that exploratory directions are vital when aggressive thresholding yields low-rank global updates. Notably, all initialization strategies except zero-padding outperform FLoRIST (90.63\%), demonstrating that principled rank discovery via ScreeNOT is the primary driver of SpecTraL's performance.
\begin{table}[t]
\centering
\caption{Ablation of initialization strategies with ScreeNOT thresholding on NICO++ (ViT-B/16, $r\!=\!32$, 75 rounds). Best and second-best results are \textbf{bolded} and \underline{underlined}.}
\label{tab:ablation_init}
\setlength{\tabcolsep}{4pt}
\renewcommand{\arraystretch}{1.1}
\small
\resizebox{\linewidth}{!}{%
\begin{tabular}{l cccccc c}
\toprule
\textbf{Initialization} & Autumn & Dim & Grass & Outdoor & Rock & Water & \cellcolor{black!8}\textbf{Avg.} \\
\midrule
Zero-padding         & 90.76 & 86.86 & 93.26 & 89.81 & 89.15 & 89.29 & \cellcolor{black!8}89.86 \\
Gaussian             & \underline{91.58} & 88.19 & 93.65 & \underline{90.86} & \underline{89.92} & \underline{90.96} & \cellcolor{black!8}\underline{90.86} \\
Trained-$A$          & 91.20 & \underline{88.91} & \underline{93.48} & 90.68 & 89.74 & 90.60 & \cellcolor{black!8}90.77 \\
Pretrained-SVD       & \textbf{91.75} & \textbf{89.27} & 93.63 & 90.62 & \textbf{90.60} & 90.62 & \cellcolor{black!8}\textbf{91.08} \\
Orthogonal compl.    & 90.81 & 88.79 & \textbf{93.97} & \textbf{90.53} & 90.19 & 90.23 & \cellcolor{black!8}90.75 \\
\bottomrule
\end{tabular}%
}
\end{table}

\subsection{Computational Heterogeneity}\label{ssec:hetero}

\begin{wrapfigure}{r}{0.45\textwidth}
    \vspace{-40pt}
    \centering
    \includegraphics[width=0.48\textwidth]{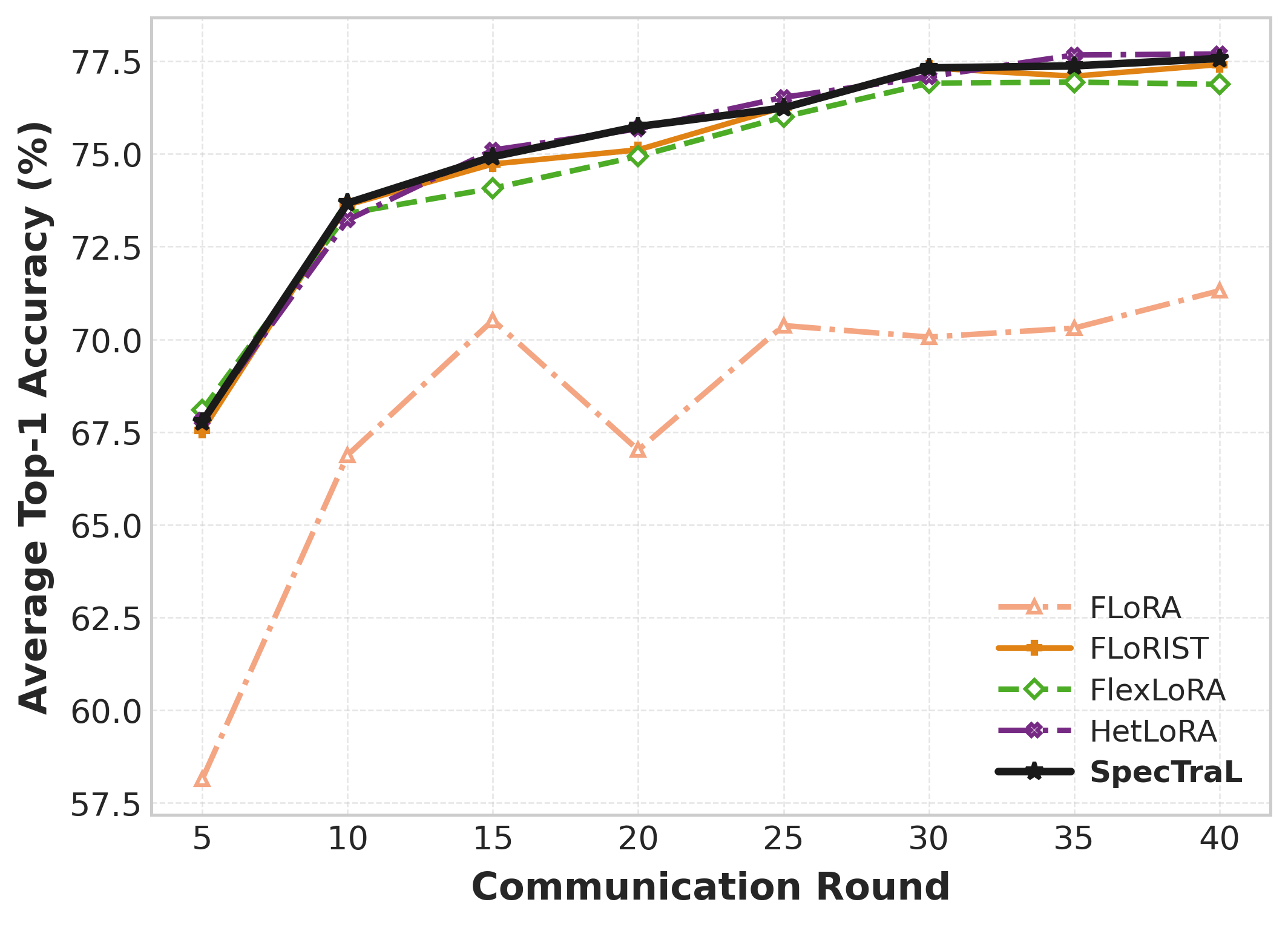}
    \caption{Convergence on DomainNet (ViT-L/16, heterogeneous ranks, 100 clients). SpecTraL matches or exceeds all baselines throughout training, while FLoRA lags due to reinitialization.}
    \label{fig:hetero_convergence}
    \vspace{-8pt}
\end{wrapfigure}

To evaluate SpecTraL under computational heterogeneity, we adopt the heavy-tail--light rank distribution of~\cite{ramesh2025florist}: among 100 clients, 40 use rank~4, 20 use rank~8, 20 use rank~16, 10 use rank~32, and 10 use rank~64. We compare against methods that natively support rank heterogeneity: HetLoRA~\cite{cho2024heterogeneous}, FLoRA, FlexLoRA, and FLoRIST.

Figure~\ref{fig:hetero_convergence} shows the convergence of average top-1 accuracy on ViT-L/16 DomainNet. SpecTraL tracks the leading methods from early rounds onward and matches or slightly exceeds HetLoRA, FLoRIST, and FlexLoRA at convergence (${\sim}77$--$78\%$). FLoRA lags throughout, reaching only ${\sim}71\%$ by round~40, a consequence of its reinitialization strategy that discards learned adapter states each round. The results confirm that SpecTraL's spectral denoising pipeline generalizes to the heterogeneous setting: even when client ranks vary by a factor of~16, principled thresholding produces global adapters that support effective local training across all capacity levels. Further gains may be achievable through heterogeneity-aware rank redistribution, which we leave as future work.







\section{Conclusion}\label{sec:conclusion}

In this work, we presented \textbf{SpecTraL}, a principled framework for federated LoRA fine-tuning that optimizes the trade-off between aggregation exactness and communication efficiency. By introducing a \textbf{QR-accelerated spectral pipeline}, we enable the server to recover the exact singular value spectrum of global updates at a fraction of the computational cost required by dense reconstruction or full SVD. Through the first application of the \textbf{ScreeNOT} estimator to this domain, SpecTraL replaces manual, heuristic energy thresholds with an automated, statistically grounded mechanism that adapts to the unique spectral signature of each transformer layer. Our extensive evaluation on vision benchmarks confirms that SpecTraL consistently achieves a superior accuracy compared to state-of-the-art baselines. Ultimately, our findings suggest that the intrinsic dimensionality of model updates is a dynamic, layer-dependent property; by treating federated aggregation as an adaptive spectral denoising problem, SpecTraL provides a robust and scalable foundation for the collaborative fine-tuning of large-scale foundation models.

\section*{Acknowledgment}
This work was supported by the NVIDIA Academic Grant Program using NVIDIA A100 (80\,GiB) GPUs accessed via the NVIDIA Brev cloud.

\bibliographystyle{unsrt} 
\bibliography{references}

\medskip


\end{document}